\documentclass[sigconf]{acmart}

\AtBeginDocument{%
  \providecommand\BibTeX{{%
    \normalfont B\kern-0.5em{\scshape i\kern-0.25em b}\kern-0.8em\TeX}}}

\usepackage{multirow}

\copyrightyear{2021}
\acmYear{2021} 
\setcopyright{iw3c2w3}
\acmConference[WWW '21]{Proceedings of the Web Conference 2021}{April 19--23, 2021}{Ljubljana, Slovenia}
\acmBooktitle{Proceedings of the Web Conference 2021 (WWW '21), April 19--23, 2021, Ljubljana, Slovenia}
\acmPrice{}
\acmDOI{10.1145/3442381.3449842}
\acmISBN{978-1-4503-8312-7/21/04}
\begin{document}

\title{TextGNN: Improving Text Encoder via Graph Neural Network in Sponsored Search}


\author{Jason Yue Zhu}
\authornote{This work was completed during the 1st author’s internship at Microsoft}
\authornote{Authors contributed equally to this work}
\affiliation{%
   \institution{Stanford University}
   \city{Stanford}
   \state{CA}
   \country{USA}}
\email{jzhu121@stanford.edu}

\author{Yanling Cui}
\authornotemark[2]
\affiliation{%
   \institution{Microsoft}
   \city{Beijing}
   \country{China}}
\email{yanling.cui@microsoft.com}

\author{Yuming Liu}
\affiliation{%
   \institution{Microsoft}
   \city{Beijing}
   \country{China}}
\email{yumliu@microsoft.com}

\author{Hao Sun}
\affiliation{%
   \institution{Microsoft}
   \city{Beijing}
   \country{China}}
\email{hasun@microsoft.com}

\author{Xue Li}
\affiliation{%
   \institution{Microsoft}
   \city{Sunnyvale}
   \state{CA}
   \country{USA}}
\email{xeli@microsoft.com}

\author{Markus Pelger}
\affiliation{%
   \institution{Stanford University}
   \city{Stanford}
   \state{CA}
   \country{USA}}
\email{mpelger@stanford.edu}

\author{Tianqi Yang}
\affiliation{%
   \institution{Microsoft}
   \city{Beijing}
   \country{China}}
\email{tianqi.yang@microsoft.com}

\author{Liangjie Zhang}
\affiliation{%
   \institution{Microsoft}
   \city{Beijing}
   \country{China}}
\email{liazha@microsoft.com}

\author{Ruofei Zhang}
\affiliation{%
   \institution{Microsoft}
   \city{Sunnyvale}
   \state{CA}
   \country{USA}}
\email{bzhang@microsoft.com}

\author{Huasha Zhao}
\affiliation{%
   \institution{Microsoft}
   \city{Sunnyvale}
   \state{CA}
   \country{USA}}
\email{huasha.zhao@microsoft.com}

\renewcommand{\shortauthors}{Zhu, et al.}

\begin{abstract}
Text encoders based on C-DSSM or transformers have demonstrated strong performance in many Natural Language Processing (NLP) tasks. Low latency variants of these models have also been developed in recent years in order to apply them in the field of sponsored search which has strict computational constraints. However these models are not the panacea to solve all the Natural Language Understanding (NLU) challenges as the pure semantic information in the data is not sufficient to fully identify the user intents. We propose the TextGNN model that naturally extends the strong twin tower structured encoders with the complementary graph information from user historical behaviors, which serves as a natural guide to help us better understand the intents and hence generate better language representations. The model inherits all the benefits of twin tower models such as C-DSSM and TwinBERT so that it can still be used in the low latency environment while achieving a significant performance gain than the strong encoder-only counterpart baseline models in both offline evaluations and online production system. In offline experiments, the model achieves a 0.14\% overall increase in ROC-AUC with a 1\% increased accuracy for long-tail low-frequency Ads, and in the online A/B testing, the model shows a 2.03\% increase in Revenue Per Mille with a 2.32\% decrease in Ad defect rate.

\end{abstract}


\keywords{Ad Relevance; Sponsored Search; Text Encoder; Graph Neural Network; Transformers; C-DSSM; BERT; Knowledge Distillation}

\maketitle

\section{Introduction}
Sponsored search refers to the business model of search engine platforms where third-party sponsored information is shown to targeted users along with other organic search results. This allows the advertisers such as manufacturers or retailers to increase the exposure of their products to more targeted potential buyers, and at the same time gives users a quicker access to solutions for their needs. Hence it has become an indispensable part of our modern web experience. While many of the existing models are very powerful for various tasks in sponsored search, there still remain three main challenges for future developments in this field: 1) while the existing models have strong performances on matching common queries with popular products, they usually still find long-tail low-frequency queries/Ads to be more challenging. The worse embedding representations in rare items are potentially caused by under-training due to naturally scarce data on these low-frequency examples. 2) while many modern models improve in implicit feature engineering on the existing input data, finding new and easily accessible data with complement information is still a promising route to greatly improve the model performance but is rarely explored. 3) the search engine systems generally have very strict constraints on computational resources and latency requirements. Many recently developed large powerful models are simply infeasible to deploy onto the highly constrained online search engine systems.

Representation learning for queries, products, or users has been a key research field with many breakthroughs over the last years and has been adopted in many production sponsored search systems \cite{Huang_2020}\cite{10.1145/3394486.3403280}\cite{10.1145/3219819.3219885}\cite{10.1145/3219819.3219897}. Convolutional Deep Structured Semantic Model (C-DSSM) \cite{shen2014learning} is among the first powerful solutions to encode text data into low-dimensional representation vectors which can be applied to downstream tasks and have efficient inference performance, but its NLU performance has been surpassed by many recently developed NLP models. The pre-trained language models emerged in recent years, such as transformers \cite{10.5555/3295222.3295349} and BERT \cite{devlin-etal-2019-bert}, have demonstrated far superior performance in many NLU tasks and even reach human level performance on many tasks. These models are better at capturing contextual information in the sentences and generate better language representation embeddings, leading to much stronger performance in downstream tasks. However, due to the complexity, these models are unfortunately not feasible to run in low latency systems without modifications. Recently, the transformer model has been modified and trained with special techniques such as knowledge distillation \cite{hinton2015distilling}, which allows us to use similar transformers structure but much smaller model called TwinBERT \cite{lu2020twinbert} to run with reasonable computational cost in the production systems while having little or no performance loss compared to the full size BERT models. This breakthrough significantly improves the user Information Retrieval experience when using search engines. However, while both C-DSSM and TwinBERT are specifically designed to be applied to the low latency systems with strong performance, they are not the panacea to fully solve all the problems in sponsored search. Their model 
ability is sometimes hindered by the limited information in the original input texts and hence still suffers in understanding many challenging low frequency inputs.

Given the strong performance of the baseline models in NLU tasks, it would be extremely difficult to further improve them solely based on the structural changes of the model without introducing new complement information. The newly developed NLP models achieve relatively small improvements with exponentially growth in model complexity, and hence reach the margin of diminishing returns making it harder to satisfy all the latency constraints. A real improved model in this field should then be able to take in additional information beyond the tradition semantic text inputs, demonstrate stronger performance over the harder low-frequency inputs, and at the same time should not significantly increase the inference time.

A natural and easily accessible data source that provides information beyond semantic text in the search engine system is users' implicit feedbacks recorded in logs in the form of clicks through the links shown to them. A click signals a connection between a query and an Ad and hence a large behavior graph based on clicks can be easily built. In the recent years, various Graph Neural Network (GNN) structures \cite{zhou2019graph} have been proposed to deal with the abundant graph-typed data and demonstrated strong performance and breakthroughs in social networks, recommendations, or natural science tasks. Motivated by the recent developments in GNN community, we are aiming to identify ways to include complementary and abundant graph-type data into the text model in a natural way. Most existing GNN models focus only on the aggregation of pre-existing neighbor features that are fixed throughout training. Instead of training the language model and the graph model separately, we want the two models to work in conjunction with each 
other to generate better query/Ad representations that can help understanding users' needs in a deeper way. 

The main contributions of this work are three-folds:
\begin{enumerate}
    \item We propose TextGNN\footnote{The BERT version implementation of the model may be found at: https://github.com/microsoft/TextGNN}, a general end-to-end framework for NLU that combines the strong language modeling text encoders with graph information processed by Graph Neural Networks to achieve stronger performance than each of its individual components.
    \item We find a systematical way to leverage graph information that greatly improves the robustness and performance by 1\% on hard examples. These samples are very challenging when only using semantic information.
    \item We trained TextGNN with knowledge distillation to get a compact model. The model has been adopted in the production system that has strict computational and latency constraints while achieving a 2.03\% increase in Revenue Per Mille with a 2.32\% decrease in Ad defect rate in the online A/B testing.
\end{enumerate}

The rest of this paper is organized as follows. Section 2 is a brief introduction of sponsored search and Ad relevance task. Section 3 reviews related literature. Section 4 discusses the details of the model, including the architecture, the construction of graph-type data, and the training methodology. Section 5 reports the experimental results of TextGNN in comparison to the baseline model under both offline and online settings with a few illustrative case study examples. Section 6 concludes the paper and briefly discusses the future directions of this work.

\section{Sponsored Search and Ad Relevance}
The TextGNN model is developed to improve the existing Ad Relevance model at a major Sponsored Search platform. In a typical sponsored search ecosystem, there are often three parties: user, advertiser and search engine platform. When the user types a query into the search engine, the goal of the platform is to understand the underlying intent of the user behind the semantic meanings of the query, and then try to best match it with a short list of Ads submitted by the advertisers alongside other organic search results. 

In the back-end when a query is received by the platform, the system will first conduct a quick but crude recall step using highly efficient Information Retrieval algorithms (such as TF-IDF \cite{Jones72astatistical} or BM25 \cite{robertson1995okapi}) to retrieve an initial list of matched candidates. The relatively long list is then passed to the downstream components for a finer filtering and final ranking using much more sophisticated but slightly less efficient models to serve the users. In both of the later steps, Deep Learning based Ad Relevance models play a key role in delivering high quality contents to the user and match advertisers' products with the potential customers.
For the Ad Relevance task, our model usually relies only on the query from a user and keywords provided by the advertiser. A \textbf{query} refers to a short text that a user typed into the search engine when he/she is looking for relevant information or product, and the model needs to identify the user's intent based on the short query. A \textbf{keyword} is a short text submitted by an advertiser that is chosen to express their intent about potential customers. The keyword is in general not visible from end users, but it is crucial for the search engine platform to match user intents.

When an Ad is displayed to a user, we call this an \textbf{impression}. The platform does not receive anything from an impression but earns revenue only when the displayed Ad is \textbf{click}ed by the user. Because of this mechanism, the search engine platform has an incentive to display the Ads that best match user intents, which directly affects the revenue. Lastly, given the scale of the traffic of the search engine, Ad Relevance models are such an indispensable component of the system and any improvement of the performance of the model can lead to huge impact on the business side of the search engine.

\section{Related Work}
\textbf{Text Encoders} including C-DSSM and Pre-trained Transformer-based Language Models (such as BERT) have achieved impressive state-of-the-art performance in many NLP tasks for their effective language or contextual word representations, hence have become one of the most important and most active research areas. 

C-DSSM is developed specifically for extracting semantic information into a low-dimension representation vector by combining convolutional layers that extract local contextual semantic information in the string with max-pooling layers that helps identifying globally important features. It is still a workhorse model used extensively in the stacks of many production search engine systems.

The large and expensive BERT model has recently become very popular. The model is usually learned in two steps. First the model is trained on extremely large corpus with unsupervised tasks such as masked language model (MLM) and next sentence prediction (NSP) to learn the general language, and then in a second step fine-tuned on the task-specific labelled data to be used in downstream tasks. Despite the strong performance of the BERT models on language representations, they are in general too expensive to be deployed in the real-time search engine systems where there are strict constraints on computation costs and latency.

\begin{figure}[tb]
\includegraphics[width=0.46\textwidth]{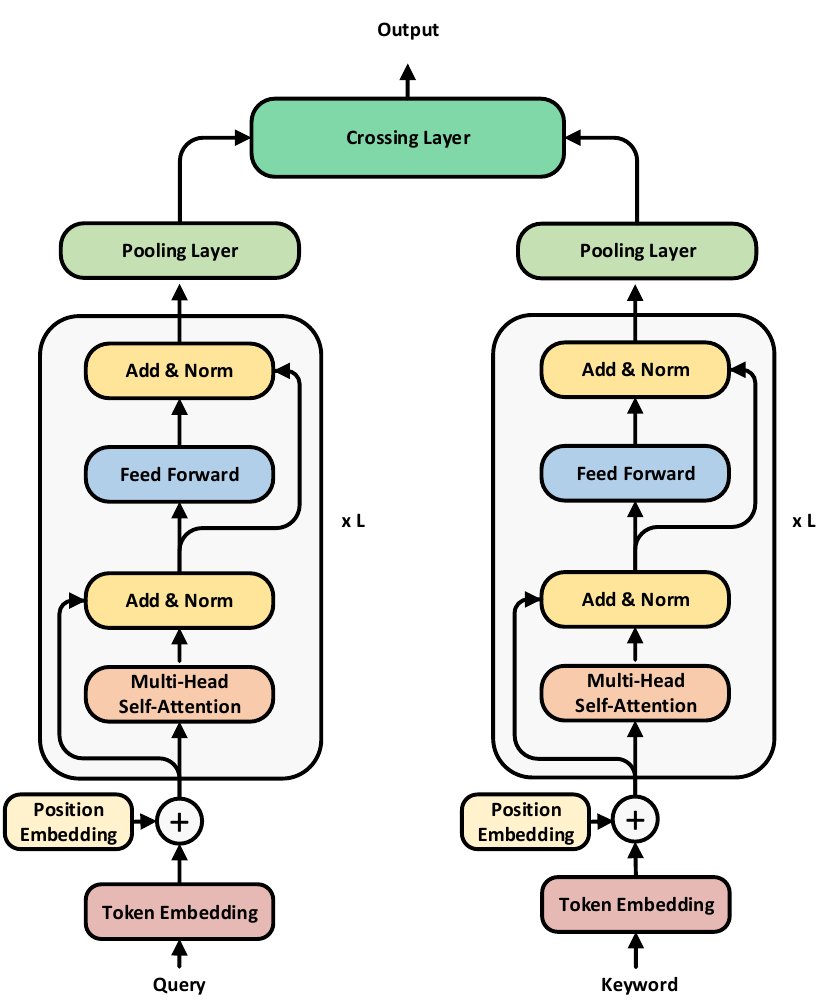}
  \caption{Architecture of the twin tower TwinBERT model}
  \label{fig:twinbert}
  \end{figure}
  
\textbf{Distilled TwinBERT} is one successful model that adapts the Transformer family models to the sponsored search applications and achieves comparable performance at reasonable inference time cost compared with heavy stacked transformer layers. The TwinBERT model as demonstrated in Figure \ref{fig:twinbert} benefits from two important techniques: 1) given two input texts, a query and a keyword, a vanilla transformer encoder would concatenate them into one input sequence, while TwinBERT has a twin tower structure to decouple the two-sentence input. Such twin tower structure is first proposed in the DSSM model \cite{huang2013learning} for web document ranking. Given that the keywords are already known to the platform, the encoded outputs of the keyword-side tower could then be pre-generated offline and fetched efficiently during inference time. Without concatenating the keyword strings, the input to the query-side tower can also be set with a low maximum length, and hence greatly reduce the inference time complexity compared to a large BERT model. 2) knowledge distillation technique is used to transfer the knowledge learnt by a teacher model to a much smaller student model. Our teacher model can be seen as a stronger version of the BM25 signal in the previous weak supervision method \cite{10.1145/3077136.3080832}. While the teacher model has strong performance, it is usually too costly and infeasible to be directly used in a production system. Knowledge distillation enables us to train a smaller model that is much faster when inference with only little or no significant loss in performance \cite{10.1145/3308558.3313466}\cite{DBLP:journals/corr/abs-1910-01108}. When a TwinBERT model with only 3 layers of encoders is used, with all the optimizations it is possible to be deployed in the real-world production systems that satisfies the strict limit from computational resources and latency requirement. 

However, as a pure language model, TwinBERT can only rely on the semantic meanings of the query-keyword pairs to infer the relationships, and in many cases when we encounter uncommon words it is still very challenging to correctly infer relevance for our main applications based on the limited input information.

\textbf{Graph Neural Network} has also become a hot research area in recent years due to its efficacy in dealing with complex graph data. Graph Convolutional Networks (GCN) \cite{Kipf:2016tc}, GraphSage \cite{NIPS2017_6703}, and Graph Attention Networks (GAT) \cite{velickovic2018graph} are among the most popular GNN models that can effectively propagate neighbor information in a graph through connected edges and hence are able to generate convincing and highly interpretable results on many graph specific tasks such as node/edge/graph property predictions. Recently there are also attempts to bring GNN to the sponsored search area such as click-through rate (CTR, ratio of the number of clicks to the number of impressions) prediction \cite{10.1145/3357384.3357951}\cite{10.1145/3357384.3357833}, but so far these attempts have only focused on using GNN to generalize the interactions among the existing fixed features. There is no strong convincing story why these features naturally form a graph and the GNN itself has no impact on the generation of the features. Alternatively people have also proposed to utilize the graph information implicitly through label-propagation to unlabeled examples\cite{10.1145/1645953.1646090}, but explicitly using the neighbor features in the model structure will be more efficient in aggregating complementary information as demonstrated in the experiments.

To the best of our knowledge, we are the first to extend various text encoders with a graph in a natural way, and co-train both text encoders and GNN parameters at the same time to achieve stronger performance in our downstream tasks.

\section{TextGNN}
\begin{figure*}
  \includegraphics[width=\textwidth]{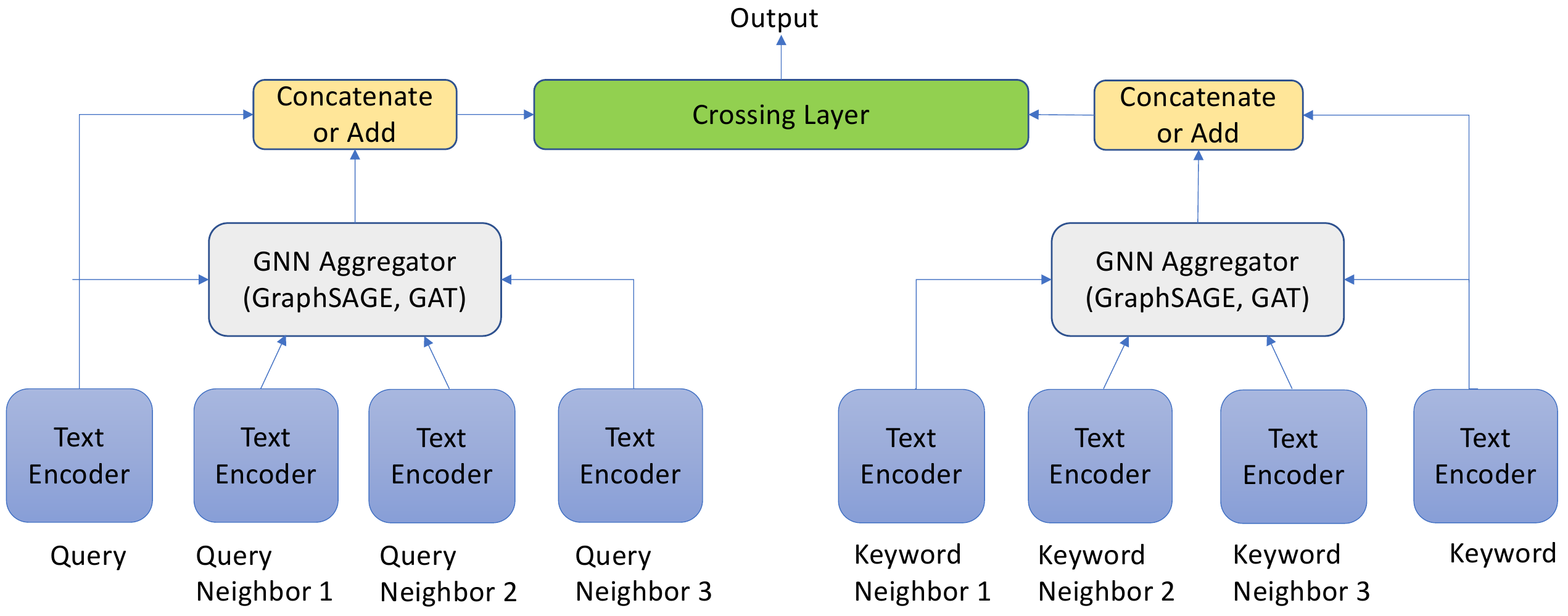}
  \caption{TextGNN Architecture: twin tower structure for decoupled generation of query/keyword embeddings}
  \label{fig:architecture}
\end{figure*}
  
In this section we will discuss the architecture of the proposed TextGNN model in Section 4.1. Then we describe the graph we used to naturally augment the semantic information of the input query-keyword sentence pairs in Section 4.2. Lastly in Section 4.3 we briefly recap knowledge distillation and its application in our model.
\subsection{Model Architecture}
The architecture of the TextGNN model is discussed in detail in this subsection and also illustrated in Figure \ref{fig:architecture}. The proposed model is a natural extension of the high-performance C-DSSM/TwinBERT baseline model with additional information from graph structured data. In sponsored search scenario, we have tens of millions candidate Ads. It is infeasible to use a complex text encoder to compute the similarity between a search query and each Ad one-by-one. Twin tower structure is a good choice for us where we could compute Ads representation vectors in advance and when a query comes, we then compute the representation vector of the query online. Notice that we only need to run the complex text encoder once for each incoming search query, compared with vanilla BERT which requires this for each unique pair. For transformer encoders, the computation cost in self-attention is also quadratic to the length of the input string. Hence, splitting the query and keyword strings for separate calculation is also much less costly than calculating the concatenated string. With these benefits in mind, our model also follows the twin tower structure of the baseline models with small encoder structure layers so that all the benefits of the twin tower structured model are inherited and hence can be deployed in the production system. Taking the query-side tower as an example, given a query and its three neighbors (defined later in the graph construction section) will all go through any general Text Encoder blocks to each generate a vector representation for the short sentence. The information from the four representation vectors is then aggregated by a GNN Aggregator to generate a single output vector. This output vector is then connected with the direct output of the text encoder of the query sentence through either concatenation or addition, similar to the idea of a Residual Connection Network \cite{he2015residual}. The combined output vector is considered as the final output of the query-side tower and can then be interacted with keyword-side output (generated from the very similar structured keyword-side tower) in the crossing layer to get the final output similar to a C-DSSM/TwinBERT model. 
\subsubsection{Text Encoder Block}
The Text Encoder block is very similar to a single tower in the C-DSSM/TwinBERT model. For example, for a transformer type text encoder, a sentence is first tokenized using the BERT WordPiece tokenizer. Trainable token embedding vectors is combined with BERT style positional embedding through addition before it go through three BERT encoder layers. The only difference with a BERT-style model is that the segment embeddings in the BERT are no longer needed as all inputs will be from the same sentence. With this structure so similar to a BERT-type one, we can conveniently load the weights from the first three layers of the pre-trained large BERT model to get a good starting point that leads to much better performance, faster model convergence, and requires significantly less training data compared to a random initialization. After the text encoder layers, we get a sequence of vectors corresponding to each token in the sentence. The vectors are then combined using a weighted-average pooling layer similar to the TwinBERT model which has demonstrated better performance in generating a single vector representation for a sentence.

The four Text Encoder blocks within a single tower are set to share the same parameters. However, the model is flexible enough to allow the two towers to have all different Text Encoder blocks, but as the TwinBERT paper shows that shared encoder blocks generally lead to slightly better performance we use that approach.

\subsubsection{GNN Aggregator}
In one tower of our TextGNN, the four text encoder blocks generate four vector representations, one for the center node (query/keyword) and the other three for its three one-hop neighbors. To aggregate the information from four vectors into one, we adopt a GNN aggregation layer, where we take the query/keyword as the central node and perform one-hop aggregation using the three neighbor nodes. The aggregation itself can be very general and use most existing GNN aggregators such as GCN, GraphSAGE, and GAT. In our experiments we found that GAT, which assigns learnable weights to the neighbors to generated a weighted average, demonstrates the strongest performance and is used in our experiments.

\subsubsection{Skip Layer}
The output vector of the query/keyword encoder is connected to the output of GNN Aggregator as the final output of the query-/keyword-side tower. This layer can be thought as a skip layer \cite{he2015residual} so that the additional GNN outputs serve as a complementary information to the text semantic representation vector. In this sense the encoder-only-models can also be considered as a special case of the TextGNN model when the GNN output is completely skipped. The two vectors are combined using either concatenation or addition. In case they have different dimensions an additional dense layer is applied after the GNN Aggregator to up/downscale the GNN output dimension to match the Text Encoder output.

\subsubsection{Crossing Layer}
Given the final outputs of the query-/keyword-side tower, the two vectors are first combined through concatenation, and then compute the similarity score using the Residual network proposed in the TwinBERT model. Formally, the residual function is defined as:
\begin{equation}
\textbf{y} = \mathcal{F}(\textbf{x}, W, b) + \textbf{x},
\end{equation}
where \textbf{x} is the concatenation of the query-side vector \textbf{q} and keyword-side vector \textbf{k} and $\mathcal{F}$ is the mapping function from \textbf{x} to the residual with parameters $W$ and $b$. A logistic regression layer is then applied to the output vector \textbf{y} to predict the binary relevance label.

\subsection{Graph Construction}
On top of the powerful structure of the model, it is also crucial to get access to high quality graph-type data. Such data should satisfy the following properties: 
\begin{enumerate}
    \item \textbf{Relevant:} since the graph neural networks propagate information along the edges, we are looking for neighbors that are highly relevant to the intent of the center node (query/keyword).
    \item \textbf{Complementary:} we expect the GNN to excel the most in situations where the language modeling part struggles to infer the intention only from the semantic meanings of the sentence, but the additional neighbors might be extremely valuable to provide complementary information that help the model to better understand the inputs. This situation happens most frequently on rare and low frequency items where the language models usually struggles on these long-tail inputs.
    \item \textbf{Accessible:} in sponsored search system, there are large amount of user input queries and candidate keywords. We try to find their neighbors in a graph. As a large graph is preferred, the neighbors need to be found with little effort and constructing the graph data should be feasible without heavy manual work, strong assumptions, or complicated structures.
\end{enumerate}

Given the requirements, we find that the user behavior graph generated from historical user clicking logs is a great candidate for our purpose. It is based on the insight that when a user inputs a query $a$ and then clicks the Ad $b$, then $b$ has to sufficiently fit the user's intent from $a$ to trigger the click. In the next two subsections, we discuss such behavior graph and its extension to address the sparse coverage issue of the behavior graph.

\begin{figure}[htb]
\includegraphics[width=0.5\textwidth]{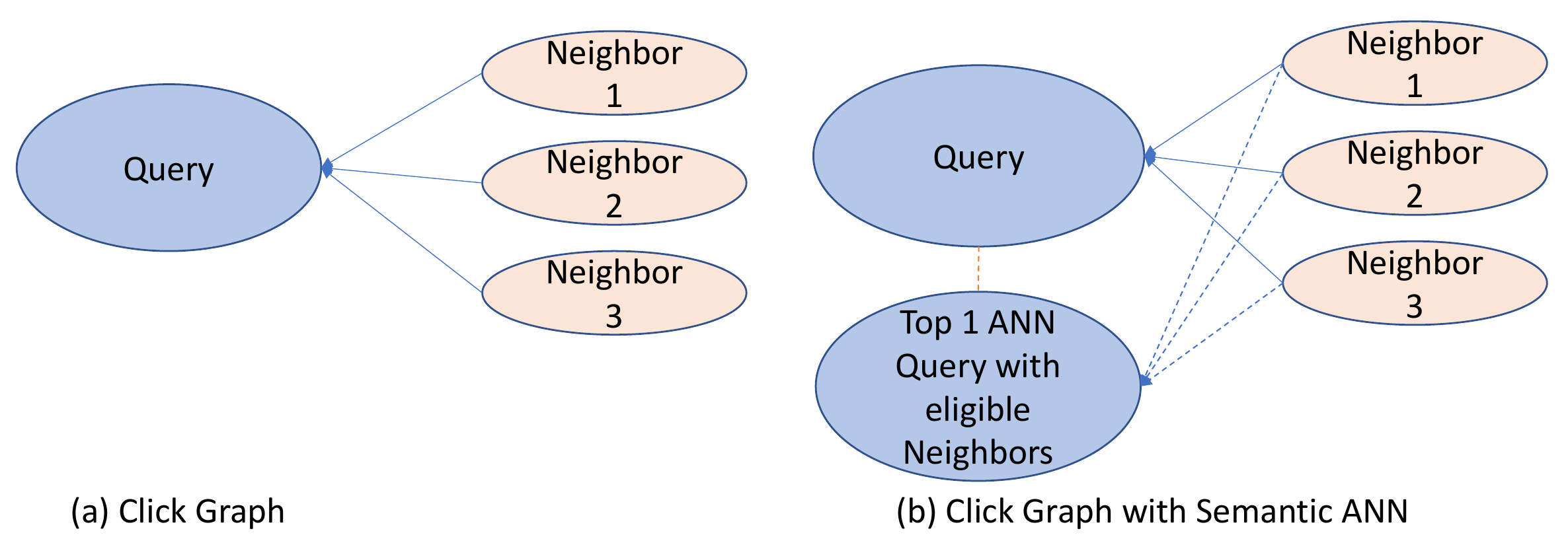}
  \caption{Click Graph Construction: use ANN proxy neighbor if no native neighbor available}
  \label{fig:neigh}
  \end{figure}
  
\subsubsection{User Click Graph}
The eligible neighbors of a query are the keyword of Ads that have been shown to be relevant to the query and received explicit positive feedback by a click. One general assumption to sort all the candidates is that the empirically observable CTR is highly correlated to the relevance between the query and the keyword. Based on this assumption, as illustrated in Figure \ref{fig:neigh}(a), we take all clicked Ads that have been shown to users at least 50 times in the past year (to partially address the issue of noisy estimates of CTR on Ads with small number of impressions) and take the top three as the neighbors.

Table \ref{tab:click_graph} shows an illustrative example, where the search query is "usps com careers login". Its top three neighbors, which are the keywords of the corresponding Ads, are listed with their historical total number of impressions and clicks. Although the first keyword "united state postal service jobs" is only shown 59 times which is significantly fewer than the third keyword "postal service hiring" with 1,721 impressions, it has a much higher CTR of 30.5\% compared to 22.3\%, indicating that users who searched for this query are more likely to find the first keyword useful, which is a strong indication of higher relevance.

\begin{table}
  \caption{Example of neighbors of a query from the Click Graph}
  \label{tab:click_graph}
  \begin{tabular}{c|c|c|c}
    \toprule
    &Clicked Neigh&Neigh&Neigh\\
    Query&Keyword&\# Impress&\# Click\\
    \midrule
     & united state & &\\
    usps com & postal service jobs & 59 & 18\\
    \cline{2-4}
    careers login & usps com employment & 344 & 92\\
    \cline{2-4}
    & postal service hiring & 1721 & 384\\
  \bottomrule
\end{tabular}
\end{table}

\subsubsection{User Click Graph with Semantic ANN}
For rare and low frequency queries/keywords, we observe by construction substantially less feedback from clicks logs. Furthermore, to avoid the noise of selecting neighbors with high CTR, we have criteria to exclude neighbors that are shown less than 50 times in the past year and this unfortunately eliminates a number of neighbors and makes the situation even worse for long-tail inputs. To address this issue, we propose a neighbor completion technique based on Approximate Nearest Neighbor (ANN) \cite{Indyk98approximatenearest} using Neighborhood Graph Search (NGS) \cite{10.1145/2393347.2393378}. As illustrated in Figure \ref{fig:neigh}(b), first we infer vector representations by a powerful C-DSSM (which is used extensively in a major sponsored search system) for all nodes in user click graph. Next, for a query that we could not identify any eligible clicked keywords, we infer its vector representation by the same C-DSSM. Then, we leverage the ANN search tool to find another query that is supposed to be semantically close enough to the original query and has the click neighbors and use its clicked keywords as approximate neighbors for the original query. This has the same spirit as the common technique of query rewriting in search engine systems but does so in a more implicit way. For keywords without any clicked queries, we find neighbors for them in a similar way.

In Table \ref{tab:click_graph_ann} we show another example that we are not able to find any eligible neighbors for the query "video games computers free", but its ANN query "no internet games" has user behavior feedback and the three approximate neighbors are obviously relevant to the original query.

\begin{table}
  \caption{Example of a query from with Semantic ANN: proxy neighbor are quite relevant to the original query}
  \label{tab:click_graph_ann}
  \begin{tabular}{c|c|c|c|c}
    \toprule
    &ANN&Clicked Neigh&Neigh&Neigh\\
    Query&Query&Keyword&\# Impress&\# Click\\
    \midrule
    video& & free games & 58&1\\
    \cline{3-5}
    games & no & online games & 260 & 4\\
    \cline{3-5}
    computers & internet &online &  & \\
    free  & games & computer games & 67 & 1\\
  \bottomrule
\end{tabular}
\end{table}

For both types of graphs, we only take at most the top three neighbors. The number of neighbors can be set as a hyper-parameter of the model framework. We choose three for following reasons: 
\begin{enumerate}
    \item More than one neighbor to provides additional complementary information while also adds robustness.
    \item Each additional neighbor means an extra run of the text encoder. Even though the encoder blocks can be run in parallel a large number of neighbors can still be computationally challenging for the system.
    \item We do not want to include more neighbors that are less relevant and introduce additional noisy information to "pollute" the encoded representation.
\end{enumerate}
Therefore, choosing three neighbors balances all the requirements and concerns.

\subsection{Knowledge Distillation}
In order to have a high performance but compact model that satisfies the computation and latency constraints, the teacher-student training framework via knowledge distillation is used. We use an expensive but high-performance RoBERTa model \cite{liu2019roberta} as the teacher model to label a very large query-keyword pair dataset, the label scores are between 0 and 1. Our model is relatively data-hungry and without this teacher model to automatically label the huge dataset, our existing human-labelled data is not sufficient to train a strong model that gets close to teacher model level performance. Since the model target, the RoBERTa score, is a continuous value, it provides more fine-grained information than the traditional binary labels. For example, a score of 0.99 indicates a stronger relevance than a score of 0.51, although both will be categorized as relevant pairs. We use mean squared error to measure the difference between the model output and the RoBERTa teacher scores.

With such a strong teacher model, we train the student TwinBERT/TextGNN model with small encoder blocks (only 3 transformer layers). Hence the student models are much more feasible in inference time but are able to achieve close to teacher model performance with only very minor performance loss. We could even further finetune the student model on a smaller human-labelled dataset with binary labels and achieve a performance surpassing the much larger teacher model. Hence, the performance of our model is not capped/limited by the teacher model.

\section{Experiments}
In this section we present experiment results of TextGNN on various tasks. We also show the comparison with the strong baseline models to show the superiority of the proposed new model and the efficacy of introducing graph information. In Section 5.1 we discuss some key statistics of the complementary graph data, and some related details of our training methods. Section 5.2 compares the performance with the baseline encoder-only models. Section 5.3 shows a more detailed sub-group analysis. Section 5.4 presents case studies of typical examples with false positive and false negative examples for TwinBERT which are correctly classified by the new TextGNN model and provide intuitive insights why the additional graph information can be valuable. Lastly in Section 5.5 we present an initial effort to apply our model to online production system and show the significant improvement over the baseline in online A/B testings.

\subsection{Data and Training Details}
For our knowledge distillation training, 397 million query-keyword pairs are scored by the teacher RoBERTa model. The student models are initialized using the parameters of the first three transformer layers of the 12-layer uncased BERT-base checkpoint \cite{Wolf2019HuggingFacesTS}. The models are evaluated on a small evaluation dataset consisting of 243 thousand human labelled samples. The query and keyword pairs were given labels with five different levels of relevance: excellent, perfect, good, fair, and bad. In the evaluation stage the first four levels excellent, perfect, good, and fair are mapped as positive samples (label 1) where the bad category is kept as negative category (label 0). The model ROC-AUC is our main metric for evaluation.

We construct the behavior click graph based on the historical search engine click logs from July 2019 to June 2020. Here in Table \ref{tab:coverage} we present some statistics on the neighbor coverage comparing the two ways of graph constructions. Here are some key observations:
\begin{enumerate}
    \item Without the added ANN neighbors, almost 2/3 of the queries miss neighbors from the user click graph. The situation is significantly better for keywords as the majority of the Ads have been shown and clicked by users.
    \item With the ANN search, we essentially increase the neighbor coverage to almost 100\%.
    \item Among all nodes, the majority of them have at least three eligible neighbors. For the examples with less than 3 neighbors, dummy padding are added.
\end{enumerate}

  
  \begin{table}
  \caption{Coverage Summary of Two Graph Construction Methods: almost full coverage after adopting ANN Neighbors}
  \label{tab:coverage}
  \begin{tabular}{l|cc|cc}
    \toprule
    & \multicolumn{2}{c|}{Click Only}&\multicolumn{2}{c}{ANN}\\
    \cline{2-3}\cline{4-5}
    &Q&K&Q&K\\
    \midrule
    1 Neighbor & 4\% & 7\% & 5\% & 7\%\\
    2 Neighbors & 3\% & 4\% & 3\% & 4\%\\
    3 Neighbors & 30\% & 76\% & 92\% & 88\%\\
    \midrule
    \textbf{Coverage} & \textbf{37\%} & \textbf{87\%}  & \textbf{100\%}  & \textbf{99\%} \\
  \bottomrule
\end{tabular}
\end{table}

\subsection{Model Performance Results}
In the experiment we train the baseline TwinBERT model and the new TextGNN model with the same common hyper-parameters for a fair comparison. The same training dataset files were used by both models, but the additional neighbor information is not read by the baseline TwinBERT model as it does not have the mechanism to process the additional information.

Tabel \ref{tab:auc} presents the ROC-AUC values of the baseline model and TextGNN based on two different types of graphs. We see that the addition of GNN has significantly improves the performance of the baseline model and the performance increase of this magnitude will lead to a huge difference in revenue for large scale systems.

  \begin{table}[h]
  \caption{ROC-AUC Comparison: TextGNN with ANN Neighbor Graph significantly outperform baseline TwinBERT}
  \label{tab:auc}
  \begin{tabular}{c|c}
    \toprule
    Model & AUC\\
    \midrule
    TwinBERT & 0.8459 \\
    TextGNN & 0.8461  \\
    TextGNN with ANN Neighbor & \textbf{0.8471}\\
  \bottomrule
\end{tabular}
\end{table}

\subsection{Sub-group Analysis}
In addition to showing the stronger overall performance of the TextGNN models over the baseline, we also conduct a more detailed sub-group analysis on inference results to confirm that the TextGNN models indeed improve on the tail examples just as expected.

We split the validation data into three bins by the Ads frequency in the dataset (as a proxy for their population frequency of impressions). 43\% of the samples are Ads that have been shown only once (among 243k samples) which are the rare examples, and 12\% of the samples have been shown twice. Even though the tail Ads individually are rarely recalled and shown to users, they consist of the majority portion of the total traffic and the improvements on these long-tail examples can lead to significant benefits.

\begin{figure}[tb]
\includegraphics[width=0.5\textwidth]{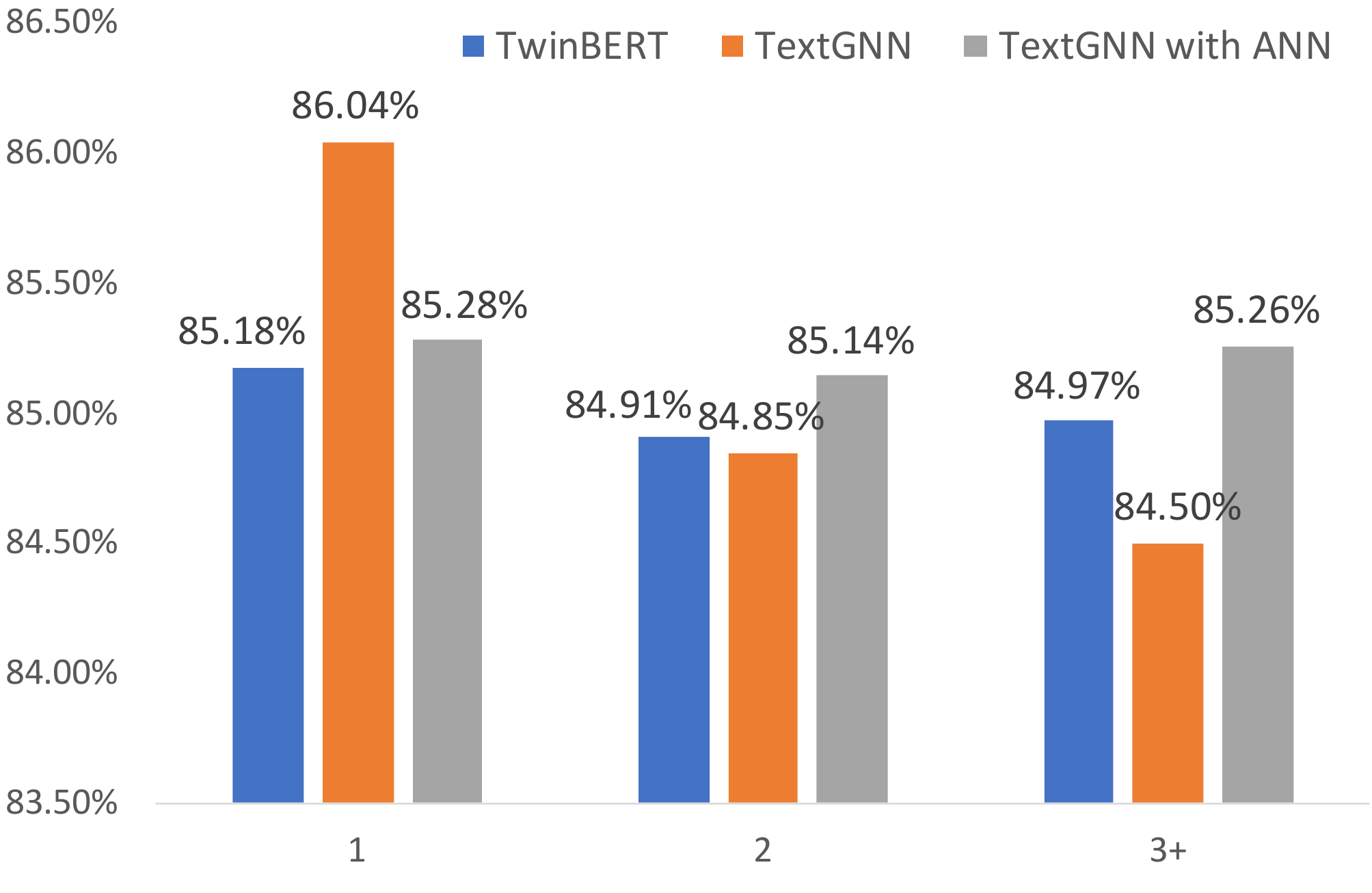}
  \caption{Performance on Different Subgroups of Data by Ads Frequency: TextGNN with vanilla click neighbor achieves extremely large gain in low frequency Ads, while the ANN version outperforms the baseline across the board}
  \label{fig:perf_by_bins}
  \end{figure}

We see the results in Figure \ref{fig:perf_by_bins} that the TextGNN model based on vanilla click graph shows an extremely large improvement in the most rare Ads, but the performance downgrades in common ones. Our hypothesis is that in the more common examples the semantic information is already good, and the limited additional information from a sparse graph is not enough to offset the potential under-fitting from a more complex model. Once we adopt ANN to generate a more complete graph, we see the TextGNN model demonstrates stronger performance than baseline across the board.

Lastly, we note that the non-ANN version is still much stronger than the ANN version in the bin of the most rare Ads, potentially because the ANN proxy neighbors are on average having lower quality than the native neighbors, and hence introduce noise to the model. This analysis also reveals a future direction to further improve the model where we can potentially use the sample frequency as a simple indicator to switch between various candidate models based on their strength within different sub-groups.

\subsection{Case Studies}
\begin{table*}[!htb]
\caption{Case study Examples: neighbors provide crucial complementary information}
  \label{tab:case_study}
\begin{tabular}{llll}
\hline
\hline
\multicolumn{4}{l}{\textbf{False Positive Examples}}                                                                                                                                    \\ \hline
Query                                        & \multicolumn{1}{l|}{Query Neighbors}                  & Keyword                                   & Keyword Neighbors                    \\ \hline
\multirow{3}{*}{achilles heel}               & \multicolumn{1}{l|}{what is an achilles heel}         & \multirow{3}{*}{plantar fasciitis shoes}  & shoes plantar fasciitis heel pain    \\
                                             & \multicolumn{1}{l|}{what is achilles heel}            &                                           & work shoes plantar fasciitis         \\
                                             & \multicolumn{1}{l|}{causes heel spurs}                &                                           & tennis shoes good plantar fasciitis  \\ \hline\hline
\multirow{3}{*}{animal repellent   products} & \multicolumn{1}{l|}{animal repeller}                  & \multirow{3}{*}{animal odor}              & best cleaning remove \& product home \\
                                             & \multicolumn{1}{l|}{keep squirrel out attic}          &                                           & air fresheners home                  \\
                                             & \multicolumn{1}{l|}{animal repellent}                 &                                           & best air fresheners                  \\ \hline\hline
\multicolumn{4}{l}{\textbf{False Negative Examples}}                                                                                                                                    \\ \hline
Query                                        & \multicolumn{1}{l|}{Query Neighbors}                  & Keyword                                   & Keyword Neighbors                    \\ \hline
\multirow{3}{*}{sharding}                    & \multicolumn{1}{l|}{mongodb cluster}                  & \multirow{3}{*}{sql server}               & sql server download windows 10       \\
                                             & \multicolumn{1}{l|}{database sharding}                &                                           & sql server hosting                   \\
                                             & \multicolumn{1}{l|}{N/A}                              &                                           & sequel server database               \\ \hline\hline
\multirow{3}{*}{use imovie}                  & \multicolumn{1}{l|}{imovies}                          & \multirow{3}{*}{adobe premiere}           & adobe premiere pro mac               \\
                                             & \multicolumn{1}{l|}{imovie 11 tutorials}              &                                           & adobe premier mac                    \\
                                             & \multicolumn{1}{l|}{imovie video editor}              &                                           & use imovie                           \\ \hline
\end{tabular}
\end{table*}

We expect the introduction of graph data to improve the model performance especially on tail inputs that are often seen as "hard" samples for the baseline models. In table \ref{tab:case_study}, we present some "hard" cases to demonstrate the value that graph data could bring.

\subsubsection{False-positive Examples of TwinBERT}
The first example shows that the user searched for the Greek methology "achilles heel", which was incorrectly determined by TwinBERT as relevant to plantar fasciitis shoes. From the semantic meaning, heel is very close to shoes and the achilles ankle is highly related to the pain of tendon. However, the neighbors strongly indicate that people who search for this query are actually looking for the story from Greek mythology and not the foot injury.

The second example shows that TwinBERT determines that "animal repellent products" is highly relevant to animal cleaning product. From the semantic meaning it is true that repellent is close in meaning to the word "remove" but the two products are used for completely different purposes. When averaging over the neighbors it is very clear that this is a negative example.

\subsubsection{False-negative Examples of TwinBERT}
The query "sharding" is a very specific concept in database systems on how large data are split and stored. Without the domain knowledge it is very hard to understand such an uncommon word. Furthermore, the word is tokenized to: [CLS], sha, \#\#rdi, \#\#ng, [SEP] by the BERT WordPiece tokenizer, making it essentially an impossible task for TwinBERT to identify the relevance. However, from the historical user behaviors we clearly see both sides taking the very important common words "database", hence allowing the TextGNN model to leverage on the user behavior to identify domain specific connections and find the hidden relevance.

The second false-negative one is an example of two video editing softwares on the Mac platform. Without the domain knowledge is it impossible to conclude from the semantic meaning that adobe premier mac is a video editing software. However, since the query string is identified as a neighbor of the keyword, our graph model can use this information to find the correct connection.

\subsection{Online A/B Test}
A slightly simplified version of our TextGNN model has already been successfully deployed in a major sponsored search platform and demonstrated significant performance gains. We have evaluated the performance of the models on the sponsored product advertising system where user search queries are matched with products with rich information provided by advertisers. In this initial effort we choose C-DSSM as the text encoder for its much faster inference time in the application of large-scale Ads corpus and use graph aggregators only on the product side of the tower. Note again that the product side representations can be generated offline in advance and hence at online service stage the latency is identical to a traditional C-DSSM model. We use the TextGNN model outputs as features to be feed into a downstream online product advertising system and evaluated the efficacy of this simple model in both offline and online settings.

For evaluation, we randomly sampled examples from online logs and labeled the data manually by human experts and observe on average 1.3\% (we only show normalized relative numbers due to business confidentiality) PR-AUC lift across different validation sets when comparing the simplified TextGNN model with the baseline C-DSSM model.

The online A/B testing results of the TextGNN model are summarized in Table \ref{tab:ab} as we applied the model to both recall and relevance stage of the Ads serving in the system, where we observe significant gains in several normalized key online metrics numbers that are crucial for our sponsored search system. The two most important metrics are:
\begin{enumerate}
    \item \textbf{Revenue Per Mille (RPM):} the revenue gained for every thousand search requests, which is one of the most important online metrics for sponsored search.
    \item \textbf{Ad Defect Rate:} the ratio of irrelevant Ad impressions with respect to total number of Ad impressions. In online A/B test, this ratio is approximated by sampling Ad impressions and submitting them for human-evaluated labels. This is highly correlated to user satisfaction and hence is considered as a very crucial metric.
\end{enumerate}

As shown in the table, the TextGNN model yields very impressive results as it can greatly boost the RPM and reduce the Ad Defect Rate, which is a strong sign that model could help to improve revenue and user experience simultaneously. It's worthy pointing out that current production model already contains many advanced sub-models and features so the magnitude of the improvement in the online KPI here is considered as a significant gain for our system at the large scale.

\begin{table}[h]
  \caption{Online A/B Testing: significant improvements in production product advertising systems}
  \label{tab:ab}
  \begin{tabular}{c|c|c}
    \toprule
    Tasks & Relative RPM & Relative Ad Defect Rate\\
    \midrule
    TextGNN Relevance & +2.03\% & -2.32\% \\
    TextGNN Selection & +1.21\% & -0.34\% \\
  \bottomrule
\end{tabular}
\end{table}

\section{Conclusion}
We present a powerful NLP model TextGNN that combines two strong model structures, text encoders and GNN, into a single end-to-end framework and shows strong performance in the task of Ad relevance. The model retains the strong natural language understanding ability from the existing powerful text encoders, while complements text encoders with additional information from graph-type data to achieve stronger performance than what could be achieved from only pure semantic information. We demonstrate with experiments that the TextGNN model show overall much stronger performance than a great baseline model based only on text encoders, and that the new model demonstrates the big gains in the most difficult task of low-frequency Ads. In our next step, the ensemble model idea could be explored to automatically mix different representation model outputs based on Ads frequency to achieve even better performance.

\bibliographystyle{ACM-Reference-Format}
\bibliography{main}


\begin{thebibliography}{27}


\ifx \showCODEN    \undefined \def \showCODEN     #1{\unskip}     \fi
\ifx \showDOI      \undefined \def \showDOI       #1{#1}\fi
\ifx \showISBNx    \undefined \def \showISBNx     #1{\unskip}     \fi
\ifx \showISBNxiii \undefined \def \showISBNxiii  #1{\unskip}     \fi
\ifx \showISSN     \undefined \def \showISSN      #1{\unskip}     \fi
\ifx \showLCCN     \undefined \def \showLCCN      #1{\unskip}     \fi
\ifx \shownote     \undefined \def \shownote      #1{#1}          \fi
\ifx \showarticletitle \undefined \def \showarticletitle #1{#1}   \fi
\ifx \showURL      \undefined \def \showURL       {\relax}        \fi
\providecommand\bibfield[2]{#2}
\providecommand\bibinfo[2]{#2}
\providecommand\natexlab[1]{#1}
\providecommand\showeprint[2][]{arXiv:#2}

\bibitem[\protect\citeauthoryear{Bai, Ordentlich, Zhang, Feng, Ratnaparkhi,
  Somvanshi, and Tjahjadi}{Bai et~al\mbox{.}}{2018}]%
        {10.1145/3219819.3219897}
\bibfield{author}{\bibinfo{person}{Xiao Bai}, \bibinfo{person}{Erik
  Ordentlich}, \bibinfo{person}{Yuanyuan Zhang}, \bibinfo{person}{Andy Feng},
  \bibinfo{person}{Adwait Ratnaparkhi}, \bibinfo{person}{Reena Somvanshi},
  {and} \bibinfo{person}{Aldi Tjahjadi}.} \bibinfo{year}{2018}\natexlab{}.
\newblock \showarticletitle{Scalable Query N-Gram Embedding for Improving
  Matching and Relevance in Sponsored Search}. In
  \bibinfo{booktitle}{\emph{Proceedings of the 24th ACM SIGKDD International
  Conference on Knowledge Discovery \& Data Mining}} (London, United Kingdom)
  \emph{(\bibinfo{series}{KDD '18})}. \bibinfo{publisher}{Association for
  Computing Machinery}, \bibinfo{address}{New York, NY, USA},
  \bibinfo{pages}{52–61}.
\newblock
\showISBNx{9781450355520}
\urldef\tempurl%
\url{https://doi.org/10.1145/3219819.3219897}
\showDOI{\tempurl}


\bibitem[\protect\citeauthoryear{Dehghani, Zamani, Severyn, Kamps, and
  Croft}{Dehghani et~al\mbox{.}}{2017}]%
        {10.1145/3077136.3080832}
\bibfield{author}{\bibinfo{person}{Mostafa Dehghani}, \bibinfo{person}{Hamed
  Zamani}, \bibinfo{person}{Aliaksei Severyn}, \bibinfo{person}{Jaap Kamps},
  {and} \bibinfo{person}{W.~Bruce Croft}.} \bibinfo{year}{2017}\natexlab{}.
\newblock \showarticletitle{Neural Ranking Models with Weak Supervision}. In
  \bibinfo{booktitle}{\emph{Proceedings of the 40th International ACM SIGIR
  Conference on Research and Development in Information Retrieval}} (Shinjuku,
  Tokyo, Japan) \emph{(\bibinfo{series}{SIGIR '17})}.
  \bibinfo{publisher}{Association for Computing Machinery},
  \bibinfo{address}{New York, NY, USA}, \bibinfo{pages}{65–74}.
\newblock
\showISBNx{9781450350228}
\urldef\tempurl%
\url{https://doi.org/10.1145/3077136.3080832}
\showDOI{\tempurl}


\bibitem[\protect\citeauthoryear{Devlin, Chang, Lee, and Toutanova}{Devlin
  et~al\mbox{.}}{2019}]%
        {devlin-etal-2019-bert}
\bibfield{author}{\bibinfo{person}{Jacob Devlin}, \bibinfo{person}{Ming-Wei
  Chang}, \bibinfo{person}{Kenton Lee}, {and} \bibinfo{person}{Kristina
  Toutanova}.} \bibinfo{year}{2019}\natexlab{}.
\newblock \showarticletitle{{BERT}: Pre-training of Deep Bidirectional
  Transformers for Language Understanding}. In
  \bibinfo{booktitle}{\emph{Proceedings of the 2019 Conference of the North
  {A}merican Chapter of the Association for Computational Linguistics: Human
  Language Technologies, Volume 1 (Long and Short Papers)}}.
  \bibinfo{publisher}{Association for Computational Linguistics},
  \bibinfo{address}{Minneapolis, Minnesota}, \bibinfo{pages}{4171--4186}.
\newblock
\urldef\tempurl%
\url{https://doi.org/10.18653/v1/N19-1423}
\showDOI{\tempurl}


\bibitem[\protect\citeauthoryear{Grbovic and Cheng}{Grbovic and Cheng}{2018}]%
        {10.1145/3219819.3219885}
\bibfield{author}{\bibinfo{person}{Mihajlo Grbovic} {and}
  \bibinfo{person}{Haibin Cheng}.} \bibinfo{year}{2018}\natexlab{}.
\newblock \showarticletitle{Real-Time Personalization Using Embeddings for
  Search Ranking at Airbnb}. In \bibinfo{booktitle}{\emph{Proceedings of the
  24th ACM SIGKDD International Conference on Knowledge Discovery \& Data
  Mining}} (London, United Kingdom) \emph{(\bibinfo{series}{KDD '18})}.
  \bibinfo{publisher}{Association for Computing Machinery},
  \bibinfo{address}{New York, NY, USA}, \bibinfo{pages}{311–320}.
\newblock
\showISBNx{9781450355520}
\urldef\tempurl%
\url{https://doi.org/10.1145/3219819.3219885}
\showDOI{\tempurl}


\bibitem[\protect\citeauthoryear{Hamilton, Ying, and Leskovec}{Hamilton
  et~al\mbox{.}}{2017}]%
        {NIPS2017_6703}
\bibfield{author}{\bibinfo{person}{Will Hamilton}, \bibinfo{person}{Zhitao
  Ying}, {and} \bibinfo{person}{Jure Leskovec}.}
  \bibinfo{year}{2017}\natexlab{}.
\newblock \showarticletitle{Inductive Representation Learning on Large Graphs}.
\newblock In \bibinfo{booktitle}{\emph{Advances in Neural Information
  Processing Systems 30}}, \bibfield{editor}{\bibinfo{person}{I.~Guyon},
  \bibinfo{person}{U.~V. Luxburg}, \bibinfo{person}{S.~Bengio},
  \bibinfo{person}{H.~Wallach}, \bibinfo{person}{R.~Fergus},
  \bibinfo{person}{S.~Vishwanathan}, {and} \bibinfo{person}{R.~Garnett}}
  (Eds.). \bibinfo{publisher}{Curran Associates, Inc.},
  \bibinfo{pages}{1024--1034}.
\newblock
\urldef\tempurl%
\url{http://papers.nips.cc/paper/6703-inductive-representation-learning-on-large-graphs.pdf}
\showURL{%
\tempurl}


\bibitem[\protect\citeauthoryear{He, Zhang, Ren, and Sun}{He
  et~al\mbox{.}}{2015}]%
        {he2015residual}
\bibfield{author}{\bibinfo{person}{Kaiming He}, \bibinfo{person}{Xiangyu
  Zhang}, \bibinfo{person}{Shaoqing Ren}, {and} \bibinfo{person}{Jian Sun}.}
  \bibinfo{year}{2015}\natexlab{}.
\newblock \bibinfo{title}{Deep Residual Learning for Image Recognition}.
\newblock
\newblock
\urldef\tempurl%
\url{http://arxiv.org/abs/1512.03385}
\showURL{%
\tempurl}
\newblock
\shownote{cite arxiv:1512.03385Comment: Tech report.}


\bibitem[\protect\citeauthoryear{Hinton, Vinyals, and Dean}{Hinton
  et~al\mbox{.}}{2015}]%
        {hinton2015distilling}
\bibfield{author}{\bibinfo{person}{Geoffrey Hinton}, \bibinfo{person}{Oriol
  Vinyals}, {and} \bibinfo{person}{Jeff Dean}.}
  \bibinfo{year}{2015}\natexlab{}.
\newblock \bibinfo{title}{Distilling the Knowledge in a Neural Network}.
\newblock
\newblock
\showeprint[arxiv]{1503.02531}~[stat.ML]


\bibitem[\protect\citeauthoryear{Huang, Sharma, Sun, Xia, Zhang, Pronin,
  Padmanabhan, Ottaviano, and Yang}{Huang et~al\mbox{.}}{2020}]%
        {Huang_2020}
\bibfield{author}{\bibinfo{person}{Jui-Ting Huang}, \bibinfo{person}{Ashish
  Sharma}, \bibinfo{person}{Shuying Sun}, \bibinfo{person}{Li Xia},
  \bibinfo{person}{David Zhang}, \bibinfo{person}{Philip Pronin},
  \bibinfo{person}{Janani Padmanabhan}, \bibinfo{person}{Giuseppe Ottaviano},
  {and} \bibinfo{person}{Linjun Yang}.} \bibinfo{year}{2020}\natexlab{}.
\newblock \showarticletitle{Embedding-based Retrieval in Facebook Search}.
\newblock \bibinfo{journal}{\emph{Proceedings of the 26th ACM SIGKDD
  International Conference on Knowledge Discovery \& Data Mining}}
  (\bibinfo{date}{Aug} \bibinfo{year}{2020}).
\newblock
\showISBNx{9781450379984}
\urldef\tempurl%
\url{https://doi.org/10.1145/3394486.3403305}
\showDOI{\tempurl}


\bibitem[\protect\citeauthoryear{Huang, He, Gao, Deng, Acero, and Heck}{Huang
  et~al\mbox{.}}{2013}]%
        {huang2013learning}
\bibfield{author}{\bibinfo{person}{Po-Sen Huang}, \bibinfo{person}{Xiaodong
  He}, \bibinfo{person}{Jianfeng Gao}, \bibinfo{person}{Li Deng},
  \bibinfo{person}{Alex Acero}, {and} \bibinfo{person}{Larry Heck}.}
  \bibinfo{year}{2013}\natexlab{}.
\newblock \showarticletitle{Learning Deep Structured Semantic Models for Web
  Search using Clickthrough Data}. \bibinfo{publisher}{ACM International
  Conference on Information and Knowledge Management (CIKM)}.
\newblock
\urldef\tempurl%
\url{https://www.microsoft.com/en-us/research/publication/learning-deep-structured-semantic-models-for-web-search-using-clickthrough-data/}
\showURL{%
\tempurl}


\bibitem[\protect\citeauthoryear{Indyk and Motwani}{Indyk and Motwani}{1998}]%
        {Indyk98approximatenearest}
\bibfield{author}{\bibinfo{person}{Piotr Indyk} {and} \bibinfo{person}{Rajeev
  Motwani}.} \bibinfo{year}{1998}\natexlab{}.
\newblock \showarticletitle{Approximate Nearest Neighbors: Towards Removing the
  Curse of Dimensionality}. \bibinfo{pages}{604--613}.
\newblock


\bibitem[\protect\citeauthoryear{Jones}{Jones}{1972}]%
        {Jones72astatistical}
\bibfield{author}{\bibinfo{person}{Karen~Spärck Jones}.}
  \bibinfo{year}{1972}\natexlab{}.
\newblock \showarticletitle{A statistical interpretation of term specificity
  and its application in retrieval}.
\newblock \bibinfo{journal}{\emph{Journal of Documentation}}
  \bibinfo{volume}{28} (\bibinfo{year}{1972}), \bibinfo{pages}{11--21}.
\newblock


\bibitem[\protect\citeauthoryear{Kim, Pantel, Duan, and Gaffney}{Kim
  et~al\mbox{.}}{2009}]%
        {10.1145/1645953.1646090}
\bibfield{author}{\bibinfo{person}{Soo-Min Kim}, \bibinfo{person}{Patrick
  Pantel}, \bibinfo{person}{Lei Duan}, {and} \bibinfo{person}{Scott Gaffney}.}
  \bibinfo{year}{2009}\natexlab{}.
\newblock \showarticletitle{Improving Web Page Classification by
  Label-Propagation over Click Graphs}. In
  \bibinfo{booktitle}{\emph{Proceedings of the 18th ACM Conference on
  Information and Knowledge Management}} (Hong Kong, China)
  \emph{(\bibinfo{series}{CIKM '09})}. \bibinfo{publisher}{Association for
  Computing Machinery}, \bibinfo{address}{New York, NY, USA},
  \bibinfo{pages}{1077–1086}.
\newblock
\showISBNx{9781605585123}
\urldef\tempurl%
\url{https://doi.org/10.1145/1645953.1646090}
\showDOI{\tempurl}


\bibitem[\protect\citeauthoryear{Kipf and Welling}{Kipf and Welling}{2017}]%
        {Kipf:2016tc}
\bibfield{author}{\bibinfo{person}{Thomas~N. Kipf} {and} \bibinfo{person}{Max
  Welling}.} \bibinfo{year}{2017}\natexlab{}.
\newblock \showarticletitle{{Semi-Supervised Classification with Graph
  Convolutional Networks}}. In \bibinfo{booktitle}{\emph{Proceedings of the 5th
  International Conference on Learning Representations}} (Palais des
  Congr{\`e}s Neptune, Toulon, France) \emph{(\bibinfo{series}{ICLR '17})}.
\newblock
\urldef\tempurl%
\url{https://openreview.net/forum?id=SJU4ayYgl}
\showURL{%
\tempurl}


\bibitem[\protect\citeauthoryear{Li, Luo, Sun, Zhang, Han, Chu, Zhang, and
  Zhang}{Li et~al\mbox{.}}{2019b}]%
        {10.1145/3308558.3313466}
\bibfield{author}{\bibinfo{person}{Xue Li}, \bibinfo{person}{Zhipeng Luo},
  \bibinfo{person}{Hao Sun}, \bibinfo{person}{Jianjin Zhang},
  \bibinfo{person}{Weihao Han}, \bibinfo{person}{Xianqi Chu},
  \bibinfo{person}{Liangjie Zhang}, {and} \bibinfo{person}{Qi Zhang}.}
  \bibinfo{year}{2019}\natexlab{b}.
\newblock \showarticletitle{Learning Fast Matching Models from Weak
  Annotations}. In \bibinfo{booktitle}{\emph{The World Wide Web Conference}}.
  \bibinfo{publisher}{Association for Computing Machinery},
  \bibinfo{pages}{2985–2991}.
\newblock


\bibitem[\protect\citeauthoryear{Li, Cui, Wu, Zhang, and Wang}{Li
  et~al\mbox{.}}{2019a}]%
        {10.1145/3357384.3357951}
\bibfield{author}{\bibinfo{person}{Zekun Li}, \bibinfo{person}{Zeyu Cui},
  \bibinfo{person}{Shu Wu}, \bibinfo{person}{Xiaoyu Zhang}, {and}
  \bibinfo{person}{Liang Wang}.} \bibinfo{year}{2019}\natexlab{a}.
\newblock \showarticletitle{Fi-GNN: Modeling Feature Interactions via Graph
  Neural Networks for CTR Prediction}. In \bibinfo{booktitle}{\emph{Proceedings
  of the 28th ACM International Conference on Information and Knowledge
  Management}} (Beijing, China) \emph{(\bibinfo{series}{CIKM '19})}.
  \bibinfo{publisher}{Association for Computing Machinery},
  \bibinfo{address}{New York, NY, USA}, \bibinfo{pages}{539–548}.
\newblock
\showISBNx{9781450369763}
\urldef\tempurl%
\url{https://doi.org/10.1145/3357384.3357951}
\showDOI{\tempurl}


\bibitem[\protect\citeauthoryear{Liu, Ott, Goyal, Du, Joshi, Chen, Levy, Lewis,
  Zettlemoyer, and Stoyanov}{Liu et~al\mbox{.}}{2019}]%
        {liu2019roberta}
\bibfield{author}{\bibinfo{person}{Yinhan Liu}, \bibinfo{person}{Myle Ott},
  \bibinfo{person}{Naman Goyal}, \bibinfo{person}{Jingfei Du},
  \bibinfo{person}{Mandar Joshi}, \bibinfo{person}{Danqi Chen},
  \bibinfo{person}{Omer Levy}, \bibinfo{person}{Mike Lewis},
  \bibinfo{person}{Luke Zettlemoyer}, {and} \bibinfo{person}{Veselin
  Stoyanov}.} \bibinfo{year}{2019}\natexlab{}.
\newblock \bibinfo{title}{RoBERTa: A Robustly Optimized BERT Pretraining
  Approach}.
\newblock
\newblock
\showeprint[arxiv]{1907.11692}~[cs.CL]


\bibitem[\protect\citeauthoryear{Lu, Jiao, and Zhang}{Lu et~al\mbox{.}}{2020}]%
        {lu2020twinbert}
\bibfield{author}{\bibinfo{person}{Wenhao Lu}, \bibinfo{person}{Jian Jiao},
  {and} \bibinfo{person}{Ruofei Zhang}.} \bibinfo{year}{2020}\natexlab{}.
\newblock \bibinfo{title}{TwinBERT: Distilling Knowledge to Twin-Structured
  BERT Models for Efficient Retrieval}.
\newblock
\newblock
\showeprint[arxiv]{2002.06275}~[cs.IR]


\bibitem[\protect\citeauthoryear{Pal, Eksombatchai, Zhou, Zhao, Rosenberg, and
  Leskovec}{Pal et~al\mbox{.}}{2020}]%
        {10.1145/3394486.3403280}
\bibfield{author}{\bibinfo{person}{Aditya Pal}, \bibinfo{person}{Chantat
  Eksombatchai}, \bibinfo{person}{Yitong Zhou}, \bibinfo{person}{Bo Zhao},
  \bibinfo{person}{Charles Rosenberg}, {and} \bibinfo{person}{Jure Leskovec}.}
  \bibinfo{year}{2020}\natexlab{}.
\newblock \showarticletitle{PinnerSage: Multi-Modal User Embedding Framework
  for Recommendations at Pinterest}. In \bibinfo{booktitle}{\emph{Proceedings
  of the 26th ACM SIGKDD International Conference on Knowledge Discovery \&
  Data Mining}} (Virtual Event, CA, USA) \emph{(\bibinfo{series}{KDD '20})}.
  \bibinfo{publisher}{Association for Computing Machinery},
  \bibinfo{address}{New York, NY, USA}, \bibinfo{pages}{2311–2320}.
\newblock
\showISBNx{9781450379984}
\urldef\tempurl%
\url{https://doi.org/10.1145/3394486.3403280}
\showDOI{\tempurl}


\bibitem[\protect\citeauthoryear{Robertson, Walker, Jones, Hancock-Beaulieu,
  and Gatford}{Robertson et~al\mbox{.}}{1995}]%
        {robertson1995okapi}
\bibfield{author}{\bibinfo{person}{Stephen Robertson}, \bibinfo{person}{S.
  Walker}, \bibinfo{person}{S. Jones}, \bibinfo{person}{M.~M.
  Hancock-Beaulieu}, {and} \bibinfo{person}{M. Gatford}.}
  \bibinfo{year}{1995}\natexlab{}.
\newblock \showarticletitle{Okapi at TREC-3}. In
  \bibinfo{booktitle}{\emph{Overview of the Third Text REtrieval Conference
  (TREC-3)} (\bibinfo{edition}{overview of the third text retrieval conference
  (trec–3)} ed.)}. \bibinfo{publisher}{Gaithersburg, MD: NIST},
  \bibinfo{pages}{109--126}.
\newblock
\urldef\tempurl%
\url{https://www.microsoft.com/en-us/research/publication/okapi-at-trec-3/}
\showURL{%
\tempurl}


\bibitem[\protect\citeauthoryear{Sanh, Debut, Chaumond, and Wolf}{Sanh
  et~al\mbox{.}}{2019}]%
        {DBLP:journals/corr/abs-1910-01108}
\bibfield{author}{\bibinfo{person}{Victor Sanh}, \bibinfo{person}{Lysandre
  Debut}, \bibinfo{person}{Julien Chaumond}, {and} \bibinfo{person}{Thomas
  Wolf}.} \bibinfo{year}{2019}\natexlab{}.
\newblock \showarticletitle{DistilBERT, a distilled version of {BERT:} smaller,
  faster, cheaper and lighter}.
\newblock \bibinfo{journal}{\emph{CoRR}}  \bibinfo{volume}{abs/1910.01108}
  (\bibinfo{year}{2019}).
\newblock
\showeprint[arxiv]{1910.01108}
\urldef\tempurl%
\url{http://arxiv.org/abs/1910.01108}
\showURL{%
\tempurl}


\bibitem[\protect\citeauthoryear{Shen, He, Gao, Deng, and Mesnil}{Shen
  et~al\mbox{.}}{2014}]%
        {shen2014learning}
\bibfield{author}{\bibinfo{person}{Yelong Shen}, \bibinfo{person}{Xiaodong He},
  \bibinfo{person}{Jianfeng Gao}, \bibinfo{person}{Li Deng}, {and}
  \bibinfo{person}{Gregoire Mesnil}.} \bibinfo{year}{2014}\natexlab{}.
\newblock \showarticletitle{Learning Semantic Representations Using
  Convolutional Neural Networks for Web Search}. \bibinfo{publisher}{WWW 2014}.
\newblock
\urldef\tempurl%
\url{https://www.microsoft.com/en-us/research/publication/learning-semantic-representations-using-convolutional-neural-networks-for-web-search/}
\showURL{%
\tempurl}


\bibitem[\protect\citeauthoryear{Vaswani, Shazeer, Parmar, Uszkoreit, Jones,
  Gomez, Kaiser, and Polosukhin}{Vaswani et~al\mbox{.}}{2017}]%
        {10.5555/3295222.3295349}
\bibfield{author}{\bibinfo{person}{Ashish Vaswani}, \bibinfo{person}{Noam
  Shazeer}, \bibinfo{person}{Niki Parmar}, \bibinfo{person}{Jakob Uszkoreit},
  \bibinfo{person}{Llion Jones}, \bibinfo{person}{Aidan~N. Gomez},
  \bibinfo{person}{undefinedukasz Kaiser}, {and} \bibinfo{person}{Illia
  Polosukhin}.} \bibinfo{year}{2017}\natexlab{}.
\newblock \showarticletitle{Attention is All You Need}. In
  \bibinfo{booktitle}{\emph{Proceedings of the 31st International Conference on
  Neural Information Processing Systems}} (Long Beach, California, USA)
  \emph{(\bibinfo{series}{NIPS'17})}. \bibinfo{publisher}{Curran Associates
  Inc.}, \bibinfo{address}{Red Hook, NY, USA}, \bibinfo{pages}{6000–6010}.
\newblock
\showISBNx{9781510860964}


\bibitem[\protect\citeauthoryear{Velickovic, Cucurull, Casanova, Romero, Liò,
  and Bengio}{Velickovic et~al\mbox{.}}{2018}]%
        {velickovic2018graph}
\bibfield{author}{\bibinfo{person}{Petar Velickovic}, \bibinfo{person}{Guillem
  Cucurull}, \bibinfo{person}{Arantxa Casanova}, \bibinfo{person}{Adriana
  Romero}, \bibinfo{person}{Pietro Liò}, {and} \bibinfo{person}{Yoshua
  Bengio}.} \bibinfo{year}{2018}\natexlab{}.
\newblock \showarticletitle{Graph Attention Networks}.
\newblock \bibinfo{journal}{\emph{ICLR}} (\bibinfo{year}{2018}).
\newblock


\bibitem[\protect\citeauthoryear{Wang and Li}{Wang and Li}{2012}]%
        {10.1145/2393347.2393378}
\bibfield{author}{\bibinfo{person}{Jingdong Wang} {and}
  \bibinfo{person}{Shipeng Li}.} \bibinfo{year}{2012}\natexlab{}.
\newblock \showarticletitle{Query-Driven Iterated Neighborhood Graph Search for
  Large Scale Indexing}. In \bibinfo{booktitle}{\emph{Proceedings of the 20th
  ACM International Conference on Multimedia}} (Nara, Japan)
  \emph{(\bibinfo{series}{MM '12})}. \bibinfo{publisher}{Association for
  Computing Machinery}, \bibinfo{address}{New York, NY, USA},
  \bibinfo{pages}{179–188}.
\newblock
\showISBNx{9781450310895}


\bibitem[\protect\citeauthoryear{Wolf, Debut, Sanh, Chaumond, Delangue, Moi,
  Cistac, Rault, Louf, Funtowicz, and Brew}{Wolf et~al\mbox{.}}{2019}]%
        {Wolf2019HuggingFacesTS}
\bibfield{author}{\bibinfo{person}{Thomas Wolf}, \bibinfo{person}{Lysandre
  Debut}, \bibinfo{person}{Victor Sanh}, \bibinfo{person}{Julien Chaumond},
  \bibinfo{person}{Clement Delangue}, \bibinfo{person}{Anthony Moi},
  \bibinfo{person}{Pierric Cistac}, \bibinfo{person}{Tim Rault},
  \bibinfo{person}{R'emi Louf}, \bibinfo{person}{Morgan Funtowicz}, {and}
  \bibinfo{person}{Jamie Brew}.} \bibinfo{year}{2019}\natexlab{}.
\newblock \showarticletitle{HuggingFace's Transformers: State-of-the-art
  Natural Language Processing}.
\newblock \bibinfo{journal}{\emph{ArXiv}}  \bibinfo{volume}{abs/1910.03771}
  (\bibinfo{year}{2019}).
\newblock


\bibitem[\protect\citeauthoryear{Yang, Deng, Tan, Tao, Zhang, Qin, and
  Ding}{Yang et~al\mbox{.}}{2019}]%
        {10.1145/3357384.3357833}
\bibfield{author}{\bibinfo{person}{Xiao Yang}, \bibinfo{person}{Tao Deng},
  \bibinfo{person}{Weihan Tan}, \bibinfo{person}{Xutian Tao},
  \bibinfo{person}{Junwei Zhang}, \bibinfo{person}{Shouke Qin}, {and}
  \bibinfo{person}{Zongyao Ding}.} \bibinfo{year}{2019}\natexlab{}.
\newblock \showarticletitle{Learning Compositional, Visual and Relational
  Representations for CTR Prediction in Sponsored Search}. In
  \bibinfo{booktitle}{\emph{Proceedings of the 28th ACM International
  Conference on Information and Knowledge Management}} (Beijing, China)
  \emph{(\bibinfo{series}{CIKM '19})}. \bibinfo{publisher}{Association for
  Computing Machinery}, \bibinfo{address}{New York, NY, USA},
  \bibinfo{pages}{2851–2859}.
\newblock
\showISBNx{9781450369763}
\urldef\tempurl%
\url{https://doi.org/10.1145/3357384.3357833}
\showDOI{\tempurl}


\bibitem[\protect\citeauthoryear{Zhou, Cui, Zhang, Yang, Liu, Wang, Li, and
  Sun}{Zhou et~al\mbox{.}}{2019}]%
        {zhou2019graph}
\bibfield{author}{\bibinfo{person}{Jie Zhou}, \bibinfo{person}{Ganqu Cui},
  \bibinfo{person}{Zhengyan Zhang}, \bibinfo{person}{Cheng Yang},
  \bibinfo{person}{Zhiyuan Liu}, \bibinfo{person}{Lifeng Wang},
  \bibinfo{person}{Changcheng Li}, {and} \bibinfo{person}{Maosong Sun}.}
  \bibinfo{year}{2019}\natexlab{}.
\newblock \bibinfo{title}{Graph Neural Networks: A Review of Methods and
  Applications}.
\newblock
\newblock
\showeprint[arxiv]{1812.08434}~[cs.LG]


\end{thebibliography}

\end{document}